\documentclass[journal]{IEEEtran}
\usepackage{graphicx}

\usepackage{cite}

\usepackage[cmex10]{amsmath}
\usepackage{hyperref}
\usepackage{url}

\usepackage{amssymb}
\usepackage{amsthm}

\usepackage{xcolor}
\usepackage{graphicx}
\hypersetup{
    colorlinks,
    linkcolor={red!50!black},
    citecolor={blue!50!black},
    urlcolor={blue!80!black}
}

\usepackage{array}

\newcommand{\mb}{\mathbf{m}}
\newcommand{\cb}{\mathbf{c}}
\newcommand{\Ub}{\mathbf{U}}

\newcommand{\pb}{\mathbf{p}}
\newcommand{\tb}{\mathbf{t}}
\newcommand{\xb}{\mathbf{x}}
\newcommand{\Xb}{\mathbf{X}}
\newcommand{\Yb}{\mathbf{Y}}

\newcommand{\diffterm}{C([S],K_m)}

\DeclareMathOperator{\tr}{tr}

\begin{document}
%
\title{Highly Scalable Tensor Factorization for Prediction of Drug-Protein Interaction Type}

\author{
    \IEEEauthorblockN{Adam Arany\thanks{\IEEEauthorrefmark{1}Adam Arany and Jaak Simm contributed equally as first authors.}\IEEEauthorrefmark{1}\IEEEauthorrefmark{2},
    Jaak Simm\IEEEauthorrefmark{1}\IEEEauthorrefmark{2},
    Pooya Zakeri\IEEEauthorrefmark{2},
    Tom Haber\IEEEauthorrefmark{3},}

    \IEEEauthorblockN{J\"org K. Wegner\IEEEauthorrefmark{4},
    Vladimir Chupakhin\IEEEauthorrefmark{4},
    Hugo Ceulemans\IEEEauthorrefmark{4},
    Yves Moreau\IEEEauthorrefmark{2}}

    \IEEEauthorblockA{\IEEEauthorrefmark{2}ESAT-STADIUS, KU Leuven, Belgium}

    \IEEEauthorblockA{\IEEEauthorrefmark{3}Hasselt University, Belgium}

    \IEEEauthorblockA{\IEEEauthorrefmark{4}Janssen Pharmaceutica, Belgium}
}

\markboth{Presented in MLCB/MLSB 2015, NIPS workshop, December~2015}%
{Highly Scalable Bayesian Tensor Factorization for Prediction of Drug-Protein Interaction Type}

\maketitle

\begin{abstract}
The understanding of the type of inhibitory interaction plays an important role in drug design.
Therefore, researchers are interested to know whether a drug has \emph{competitive} or \emph{non-competitive} interaction to particular protein targets.
\\
\underline{Method:} to analyze the interaction types we propose factorization method \emph{Macau} which allows us to combine different measurement types into a single \emph{tensor} together with proteins and compounds.
The compounds are characterized by high dimensional 2D ECFP fingerprints.
The novelty of the proposed method is that using a specially designed noise injection MCMC sampler it can incorporate high dimensional side information, i.e., millions of unique 2D ECFP compound features, even for large scale datasets of millions of compounds.
Without the side information, in this case, the tensor factorization would be practically futile.
\\
\underline{Results:} using public IC50 and Ki data from ChEMBL we trained a model from where we can identify the latent subspace separating the two measurement types (IC50 and Ki).
The results suggest the proposed method can detect the competitive inhibitory activity between compounds and proteins.
\end{abstract}


\IEEEpeerreviewmaketitle

\section{Introduction}
In pharmaceutical research two properties, \emph{affinity} and \emph{potency}, are commonly measured \emph{in-vitro}. \emph{Affinity} measures the concentration of the tested substance needed to occupy a given portion of the available binding sites, usually measured by the inhibitory constant $K_i$ which have an intuitive meaning: the free substance concentration in chemical equilibrium where the half of the available binding sites are occupied. This value is independent of the concentration of the protein.

\emph{Potency} measures the concentration of the substance to reach a given level of effect. In case of inhibitory compounds most often measured by relative $IC_{50}$. Relative $IC_{50}$ is the concentration of the substance, where the halfway activity between the baseline and the maximal inhibition is reached.

\begin{figure}[!t]
\centering
\includegraphics[width=3.2in]{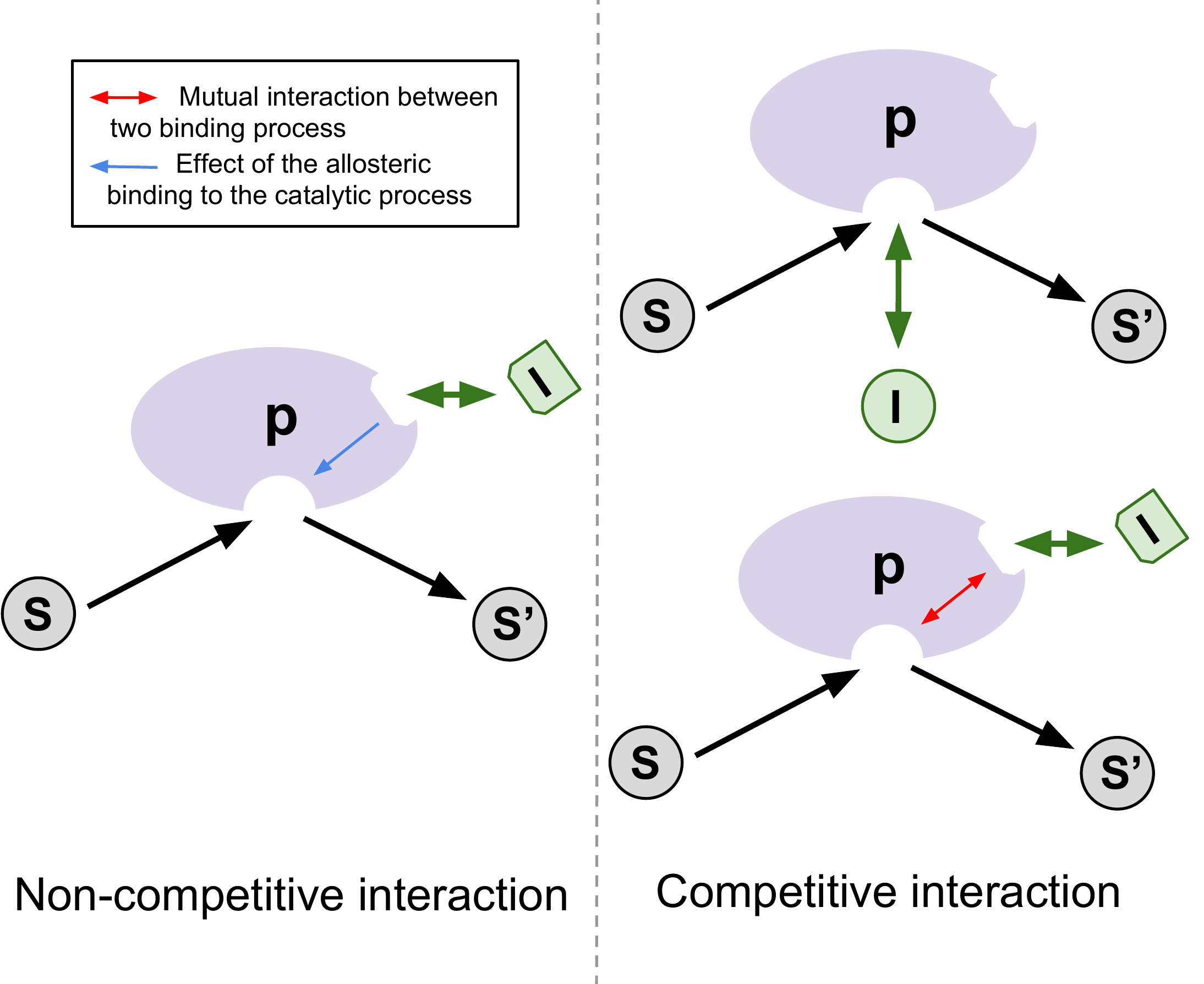}
\caption{Two main types of inhibition.}
\label{fig_inhibitors}
\end{figure}

Assuming Michaelis-Menten kinetics, it is easy to see that if there is no interaction between the binding processes, $IC_{50} = K_i$ even in the case of partial inhibitors. This can be true when there is only one substance binding to the binding site of interest, so the binding site of the inhibitor is \emph{allosteric} to the binding site where the effect is measured, illustrated on the left panel of Fig.~\ref{fig_inhibitors}. 
If, however, there is competition between the inhibitor and an endogenous ligand (illustrated on the right panel of Fig.~\ref{fig_inhibitors}), $IC_{50}$ will be dependent of the concentrations of the endogenous ligand and its interaction with the protein. The relation in the logarithmic scale has the form \cite{yung1973relationship}:
\begin{equation}
\label{eq:chengprusoff}
pK_i = pIC_{50} + C\left([S],K_m\right),
\end{equation}
where $pK_i = -\log_{10} K_i$, $pIC_{50} = -\log_{10} IC_{50}$, $[S]$ is the concentration of the endogenous substrate $S$, $K_m$ is the Michaelis constant of the enzyme with respect to $S$, and $C([S],K_m)$ is an additive constant with respect to $pIC_{50}$, depends on the experimental conditions.
This form holds in the case of uncompetitive inhibition as well.

We want to build a Bayesian model for the Eq.~\ref{eq:chengprusoff} by jointly modeling both $pIC_{50}$ and $pK_i$, using data from different proteins and compounds.
The joint model can capture whether the deviation term $C([S],K_m)$ is zero (corresponding to non-competitive interaction) or non-zero (corresponding to competitive interaction).
Therefore, we propose a model that uses low-rank approximation of the 3-way tensor $\Yb$ where $Y_{ijk}$ is the measured inhibition of protein $j$ with compound $i$ for the given inhibition measure $k$ (either $pK_i$ or $pIC_{50}$):
\begin{equation}
  \label{eq:Y-tensor}
  Y_{ijk} \approx \mathbf{1}^\top (\cb_i \circ \pb_j \circ \tb_k),
\end{equation}
where $\cb_i, \pb_j, \tb_k \in \mathbb{R}^D$ are the respective $D$-dimensional latent vectors and $\mathbf{1}$ is a vector of ones.
The tensor $\Yb$ is thin as there are only two measure types.

It is a natural property of tensor factorization that additive and uncorrelated effects will occupy different latent dimensions.
This is the case for the Eq.~\ref{eq:chengprusoff} and, thus, we expect few specific latent dimensions to capture the difference term $C([S], K_m)$.

\subsection{Side Information}
\begin{figure}[!t]
\centering
\includegraphics[width=2.3in]{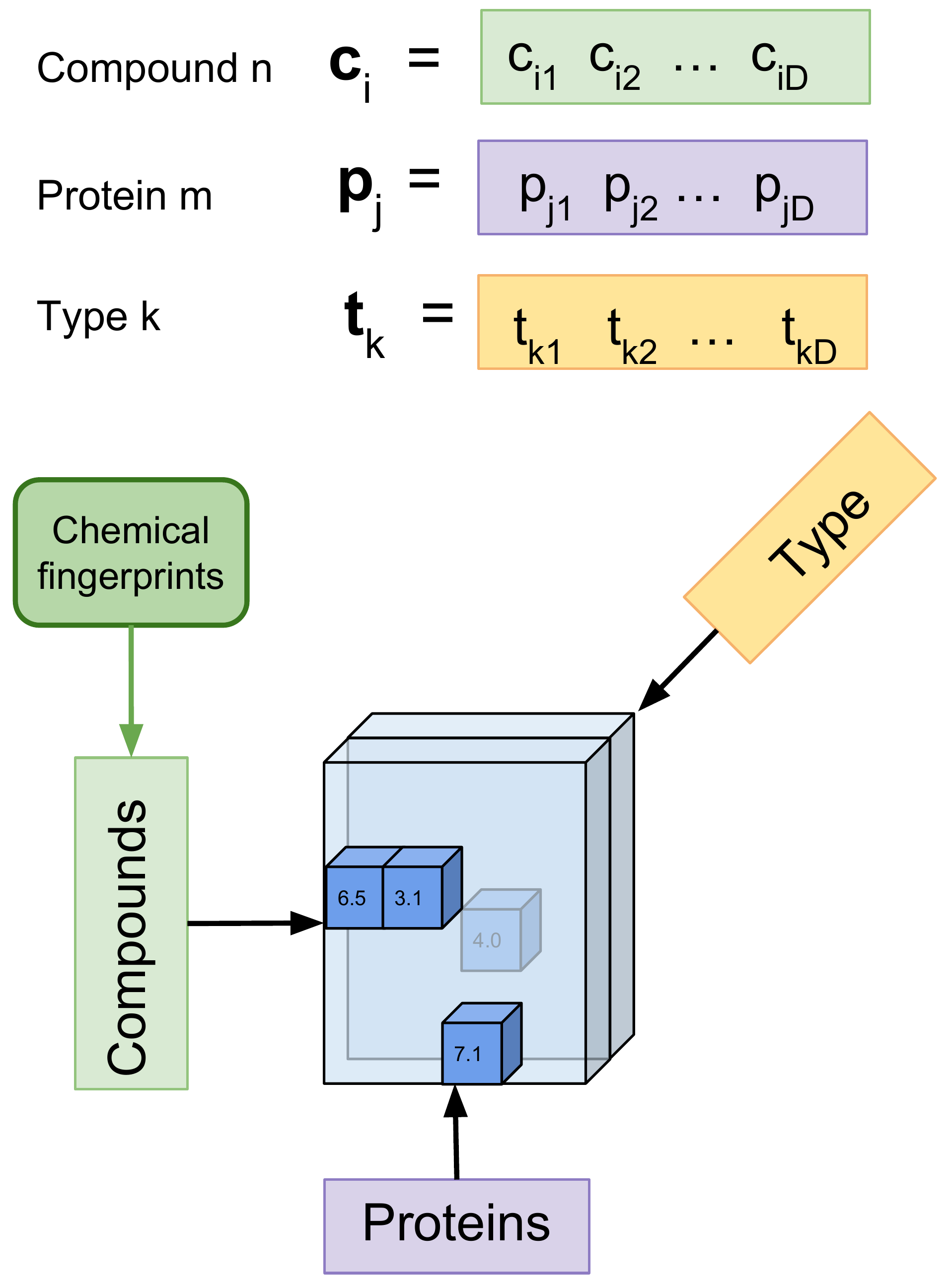}
\caption{Tensor factorization for $pIC_{50}$ and $pK_i$ modeling with side information on compounds.}
\label{fig_tensor}
\end{figure}

However, one important issue with the factorization \eqref{eq:Y-tensor} is that the tensor $\Yb$ is very sparsely observed with less than 1\% measured values.
To solve this issue we propose to use 2D ECFP chemical features as a side information for compounds while performing the factorization~\eqref{eq:Y-tensor}.
The ECFP features represent the substructures of the compound, which are strongly related to its interaction properties.
The proposed model learns a link matrix $\beta_c \in \mathbb{R}^{D \times F_c}$ to predict latent vectors $\cb_i$ from features $\xb_i \in \mathbb{R}^{F_c}$, where $F_c$ is the number of side information features.
As will become evident from the experiments the incorporation of ECFP features gives a very strong improvement to the accuracy of the model.
The general setup for the model is depicted in Fig.~\ref{fig_tensor}.

To sample from this model we develop a Bayesian factorization approach \emph{Macau} based on Gibbs sampling.
The \emph{novelty} of Macau is that in can \emph{scale} well with respect to the number of features $F_c$.
This is crucial for the current application as the dimension of the ECFP features $F_c$ is more than 100,000.
In contrast, the recent sampling-based Bayesian factorization methods Bayesian Matrix Factorization with Side Information (BMFSI)~\cite{porteous2010bayesian} and Bayesian Canonical PARAFAC with Features and Networks (BCPFN)~\cite{rai2015leveraging} require explicitly computing the covariance matrix of size $F_c \times F_c$ for linking the features to the factorization.
Additionally, to sample the link vector Cholesky decomposition is necessary, taking at least $O({F_c}^3)$, which makes these methods intractable for high-dimensional feature spaces.

Instead, Macau samples the link matrix $\beta_c$ without \emph{explicitly} computing its covariance matrix by using a specially designed noise injection sampler that we describe in the Section~\ref{sec:noise-injection}.

\section{Statistical Model}
In this section we first describe the statistical model, and explain related work in tensor factorization literature.

\subsection{General Setup}
The observation noise is assumed to be Gaussian, as was proposed in Bayesian Probabilistic Matrix Factorization~\cite{salakhutdinov2008bayesian},
\begin{equation}
  p(\Yb | \cb, \pb, \tb, \alpha) =
    \prod_{(i,j,k) \in I}
    \mathcal{N}(Y_{ijk} | \mathbf{1}^\top (\cb_i \circ \pb_j \circ \tb_k), \alpha^{-1}),
\end{equation}
where $\mathcal{N}$ is the normal distribution, $I$ is the set of tensor cells whose value has been observed and $\alpha$ is the noise precision.

Next we place a common multivariate Gaussian prior on the latent vectors of compounds
\begin{equation}
  \label{eq:latent-prior}
  p(\cb_i | \Lambda_c, \mu_c, \xb_i, \beta_c) =
    \mathcal{N}(\cb_i | \mu_c + \beta_c \xb_i, \Lambda_c),
\end{equation}
where $\mu_c \in \mathbb{R}^D$ and $\Lambda_c \in \mathbb{R}^{D \times D}$ are the shared intercept vector and precision matrix for all compounds.
The term $\beta_c \xb_i$ allows the prior mean of $\cb_i$ to depend on the compound features $\xb_i$ and, therefore, have informed predictions also for compounds with very few or no measurements.
We use analogous prior also for the latent vectors of proteins $\pb_j$ and measurements $\tb_k$.
For proteins we use basic sequence based features, described in more detail Section~\ref{sec:data}.
Without the $\beta_c \xb_i$ in the prior~\eqref{eq:latent-prior} our model is equivalent to Bayesian Probabilistic Tensor Factorization~\cite{xiong2010temporal}.
Originally, the adjustment of the prior mean was first proposed by Agarwal et al.~\cite{agarwal2009regression} for non-Bayesian models.
Recently, this side information adjustment was used for sparse tensor factorization in BCPFN~\cite{rai2015leveraging}.

\subsection{Scale Invariant Prior on the Link Matrix}
The standard approach in the previous works, for example used by Rai et al.~\cite{rai2015leveraging}, has been to add a zero-mean Gaussian prior on $\beta_c$, i.e., $\text{vec}(\beta_c) \sim \mathcal{N}(\mathbf{0}, \lambda_{\beta_c} \mathbf{I})$, where $\mathbf{I}$ is the identity matrix and $\lambda_{\beta_c} \in \mathbb{R}_+$ is precision value.

Instead, we propose a prior whose scale depends on $\Lambda_c$, making the strength of the prior \emph{invariant} to the scale of the latent variables:
\begin{equation}
  \label{eq:link-prior}
  p(\beta_c|\Lambda_c, \lambda_{\beta_c}) =
    \mathcal{N}(\text{vec}(\beta_c) | \mathbf{0}, \Lambda_c \otimes \lambda_{\beta_c}\mathbf{I}_{F_c}),
\end{equation}
where $\otimes$ is Kroenecker product.
The prior has advantage of being well suited for sampling, as we show in the next section.

\section{Gibbs Sampler}
We develop a block Gibbs sampler for the model.
For most variables the samplers are straight-forward, and due to space restrictions are left to an extended version of the paper~\cite{simm2015macau}.
The crucial question is how to sample $\beta_c$ whose dimensions in our experiments are $30 \times \text{105,672}$ but can be even larger for bigger datasets.

The main idea of our noise injection sampler is that we will construct a specific linear system whose solution corresponds to a sample from the conditional probability of $\beta_c$.
For large feature dimension $F_c$ the linear system of size $F_c \times F_c$ will be solved without explicitly computing it by using \emph{iterative solvers}, which only require (sparse) matrix multiplication operations.

\subsection{Noise Injection Sampler for the Link Matrix}
\label{sec:noise-injection}
The conditional posterior of $\beta_c$, used for Gibbs sampler, combines \eqref{eq:latent-prior} and \eqref{eq:link-prior}. Due to scale invariance of \eqref{eq:link-prior} we get
\begin{equation}
  p(\beta_c) \propto
    \exp(
      -\dfrac{1}{2}
       \tr[ ((\Ub - \Xb \beta_c^\top)^\top (\Ub - \Xb \beta_c^\top) + \lambda_{\beta_c}\beta_c^\top \beta_c) \Lambda_c ]
    )
\end{equation}
where for conciseness we have dropped the conditional terms after $\beta_c$ and use shorthands $\Ub = [\cb_1 - \mu_c, \ldots, \cb_{N_c} - \mu_c]^\top$ and $\Xb = [\xb_1, \ldots, \xb_{N_c}]^\top$.
From this we can derive the Gaussian mean and precision of $\beta_c$ as
\begin{equation}
  \label{eq:link-gibbs}
  p(\beta_c) = \mathcal{N}(\mathbf{K}^{-1} \Xb^\top \Ub, \Lambda^{-1}_c \otimes \mathbf{K}^{-1} ),
\end{equation}
where $\mathbf{K} = \Xb^\top \Xb + \lambda_{\beta_c}\mathbf{I}$.
Naively sampling from Gaussian \eqref{eq:link-gibbs} would cost $O({F_c}^3 D^3)$ time.
However, due to the special structure of the distribution \eqref{eq:link-gibbs} we can generate its sample by solving following $F_c \times F_c$ linear systems
\begin{equation}
  \label{eq:link-sampler}
  \mathbf{K} \tilde{\beta}_c = \Xb^\top(\Ub+\mathbf{E}_1) + \sqrt{\lambda_{\beta_c}} \mathbf{E}_2
\end{equation}
with respect to $\tilde{\beta}_c$, where $\mathbf{E}_1$ and $\mathbf{E}_2$ are \emph{noise} matrices where each row is drawn from $\mathcal{N}(\mathbf{0}, \Lambda_c^{-1})$. We call \eqref{eq:link-sampler} the \emph{noise injection sampler} as it adds noise to the right-hand sides of the linear systems.
The correctness of the noise injection sampler is proved in extended version, see~\cite{simm2015macau}.
The RHS of the system~\eqref{eq:link-sampler} has dimension $F_c \times D$, \emph{i.e.}, there are total of $D$ systems, which can be solved independently, using embarrassingly parallel approach.

We propose to use \emph{conjugate gradient} to solve \eqref{eq:link-sampler}.
In its iterations conjugate gradient only requires multiplication of $\mathbf{K}$ with a dense vector.
This allows us to handle side information with millions of sparse features $\Xb$ because the multiplication $\mathbf{K}\mathbf{v}$ can be expressed as two sparse-matrix-vector-products and an addition:
\begin{equation}
  \mathbf{K} \mathbf{v} = \Xb^\top (\Xb \mathbf{v}) + \lambda_{\beta_c} \mathbf{v}.
\end{equation}

In an analogous way, we introduced side information for the proteins.

The software implementation of Macau is freely available\footnote{\url{https://github.com/jaak-s/BayesianDataFusion.jl}}.

\section{Experiments}
\subsection{Data}
\label{sec:data}
The inhibitory activities were obtained from ChEMBL \cite{bento2014chembl} version 19. The dimensions of the tensor $\Yb$ are $ N_c=\text{15,073}$, $N_p=346$ and $N_t=2$, with 59,280 observed values for $pIC_{50}$, and 5,121 for $pK_i$.

We generate ECFP features of radius 3 for the compounds by using RDKit software\cite{rdkit}.
These features are sparse and high dimensional with $F_{c}=\text{105,672}$.
 
To have a basic description for proteins we use protein features based on information extracted directly from position-specific frequency matrices (PSFM). Accordingly, we link each protein to a 20 dimensional feature vector created by summing up each column in the PSFM profile as done in \cite{zakeri2014protein}.

\subsection{Results}
In this subsection we present empirical results on ChEMBL dataset. In all experiments we used Macau with $D=30$ latent dimensions.

\subsubsection{Effect of side information}
We compare Macau with and without side information in a hold-out validation setting, which we repeat 10 times. On each run, we randomly select 20\% of the $pIC_{50}$ measurements as the test set.
The mean and standard deviation of the RMSEs show that the side information plays a crucial role to have accurate predictions for $pIC_{50}$, see Table~\ref{table:MacauRMSE}.

\begin{table}[!t]
\renewcommand{\arraystretch}{1.3}
\caption{Effect of side information on RMSE (on test set)}
\label{table:MacauRMSE}
\centering
\begin{tabular}{|c||c|}
\hline
Method & RMSE (Stdev)\\
\hline
Macau with SI & \textbf{0.5970 (0.0054)}\\
Macau without SI & 0.8703 (0.0085)\\
\hline
\end{tabular}
\end{table}

\subsubsection{Latent dimensions capturing the difference term}
\begin{figure}[!t]
\centering
\includegraphics[width=3.0in]{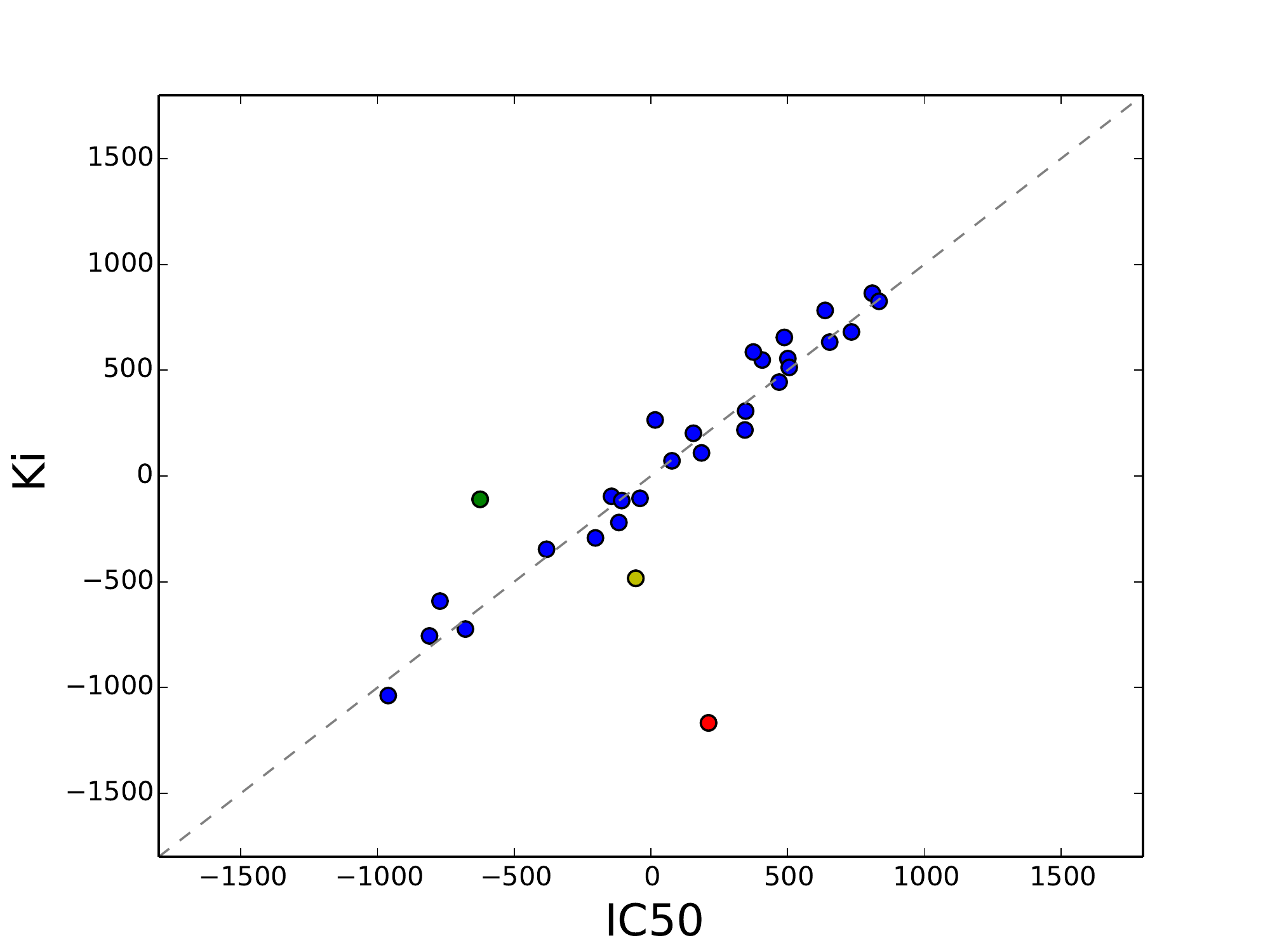}
\caption{Normalized latent dimensions of measurement types $pIC_{50}$ and $pK_i$.}
\label{fig_latent}
\end{figure}

We compare the latent vectors of $pIC_{50}$ and $pK_i$, i.e. $\tb_1$ and $\tb_2$.
Figure~\ref{fig_latent} depicts the normalized\footnote{For normalization both $t_{1,d}$ and $t_{2,d}$ are multiplied by Euclidean norms of compounds and proteins $\| \cb^{(d)} \| \| \pb^{(d)} \|$.} values of latent dimensions in the two vectors $\tb_1$ and $\tb_2$, from a single posterior sample.
As we can see from the figure, in three latent dimensions, colored green, red and yellow, the values corresponding to $pIC_{50}$ and $pK_i$ differ significantly.
The difference term $\diffterm$ in \eqref{eq:chengprusoff} is captured by these three latent dimensions, as expected.

As can be seen from the animation\footnote{Animation is available at \url{https://youtu.be/RSrVfUwYYw4}}, the behavior of these 3 latent dimensions through the posterior samples is stable and is qualitatively identical to Figure~\ref{fig_latent}.
Next we investigate the predictive properties of these 3 latent dimensions.

\subsubsection{Prediction of dominant interaction types for proteins}
We used the identified three latent dimensions from the previous section to rank the proteins according to their predicted binding behavior.
Let $\mb$ be an indicator vector where $d$-th element is 1 exactly if the $d$-th latent vector is one of the 3 selected.
We computed the absolute difference between the predicted $pIC_{50}$ and $pK_i$ values
\begin{equation*}
C_{ij} = |\mathbf{m}^\top (\cb_i \circ \pb_j \circ (\tb_2 - \tb_1))|,
\end{equation*}
 where $\tb_2$ is the latent vector for $pK_i$, and $\tb_1$ is for $pIC_{50}$. Since our model is Bayesian, we average $C_{ij}$ over all posterior samples, as in the previous sections, and denote it by $\hat{C}_{ij}$.
 
 Next, we computed the 0.95 quantile of $\hat{C}_{ij}$ for each protein $j$, which is a robust estimate for the maximal difference term $C([S],K_m)$ in~\eqref{eq:chengprusoff} for protein $j$.
 We order all the proteins according to this quantile, and extracted the top and the bottom 10 of this list.
We expect that the top 10 proteins, shown in Table~\ref{table:protein_competitive}, to have strong competitive interactions. As can be seen from the table, all 10 are enzymes, which makes it probable that our expectation is correct, because enzymes usually have some strongly competitive inhibitors.
Secondly, we expect the bottom 10 proteins, shown in Table~\ref{table:protein_noncompetitive}, to be generally non-competitive, which also seems to be the case because there are 7 receptors and 2 ion channels.
To understand what $pIC_{50}$ measures for receptors, one has to know how the assays are designed.
In the assays of receptors and ion channels, the inhibitory response is measured on the intracellular end of the signal, which is non-competitive, even though on the extracellular site there could be competition for the site.
An alternative explanation is that the proposed model detects deviations from Michaelis-Menten kinetics.

As the tables were constructed only using the selected 3 latent dimensions, it shows that the tensor factorization has the ability to extract the differential term $C([S],K_m)$ into few latent dimensions.
 
\subsubsection{Prediction of compound-protein interaction types}
In this experiment we try a challenging task to predict if a compound-protein pair have competitive interaction ($pK_i \neq pIC_{50}$).
We select 30 compound-protein pairs to the test set, for which both $pIC_{50}$ and $pK_i$ are available. 10 pairs are randomly selected from the top 100 pairs having the highest positive difference $pK_i - pIC_{50}$, next 10 from the 100 pairs having the highest negative difference, and the final 10  from the 100 pairs having the least absolute difference.
From the Macau predictions (using all latent dimensions) we compute the difference $pK_i - pIC_{50}$ for the pairs in the test set.

Macau is able to capture signal from the data as the mean predicted absolute difference, $|pK_i - pIC_{50}|$, for the competitive pairs is 0.671 and for the non-competitive pairs is 0.234. The difference is statistically significant at 1\% level (with t-test p-value of 0.0047).
\begin{table}[!t]
\renewcommand{\arraystretch}{1.3}
\caption{TOP10 Proteins with predicted competitive interaction}
\label{table:protein_competitive}
\centering
\begin{tabular}{|c||c|}
\hline
ChEMBL ID & Protein name\\
\hline
CHEMBL284 & Dipeptidyl peptidase IV\\
CHEMBL325 & Histone deacetylase 1\\
CHEMBL260 & MAP kinase p38 alpha\\
CHEMBL1865 & Histone deacetylase 6\\
CHEMBL1937 & Histone deacetylase 2\\
CHEMBL289 & Cytochrome P450 2D6\\
CHEMBL4005 & PI3-kinase p110-alpha subunit\\
CHEMBL1978 & Cytochrome P450 19A1\\
CHEMBL2581 & Cathepsin D\\
CHEMBL4793 & Dipeptidyl peptidase IX\\
\hline
\end{tabular}
\end{table}

\begin{table}[!t]
\renewcommand{\arraystretch}{1.3}
\caption{TOP10 Proteins with predicted non-competitive interaction}
\label{table:protein_noncompetitive}
\centering
\begin{tabular}{|c||c|}
\hline
ChEMBL ID & Protein name\\
\hline
CHEMBL240 & HERG\\    
CHEMBL3772 & Metabotropic glutamate receptor 1\\
CHEMBL5145 & Serine/threonine-protein kinase B-raf\\ 
CHEMBL3663 & Growth factor receptor-bound protein 2\\
CHEMBL4641 & Voltage-gated T-type calcium channel alpha-1G subunit\\
CHEMBL3230 & Sphingosine 1-phosphate receptor Edg-6\\
CHEMBL2001 & Purinergic receptor P2Y12\\    
CHEMBL1785 & Endothelin receptor ET-B\\    
CHEMBL287 & Sigma opioid receptor\\
CHEMBL3227 & Metabotropic glutamate receptor 5\\
\hline
\end{tabular}
\end{table}


\section{Conclusion}
In this paper we presented a novel scalable Bayesian tensor factorization method with side information in a novel application for the prediction of drug-protein interaction types.

We showed that the method can capture the difference between binding affinity and potency from highly sparse data.
With this model we can predict the dominant interactions types of the proteins and the competitiveness of the inhibition between compound-protein pairs.

The future work is to execute this factorization model on large scale industrial dataset and make it directly usable for drug design process.
We also plan to extend the model for slow tight-binding inhibitors.


\section*{Acknowledgment}
Authors thank Thierry Louat for insightful discussions.
Jaak Simm, Adam Arany, Pooya Zakeri and Yves Moreau are funded by Research Council KU Leuven (CoE PFV/10/016 SymBioSys) and by Flemish Government (IOF, Hercules Stitching, iMinds Medical Information Technologies SBO 2015, IWT: O\&O ExaScience Life Pharma; ChemBioBridge, PhD grants).

\bibliography{ref}
\bibliographystyle{IEEEtran}
\end{document}